\pdfoutput=1

\documentclass[11pt]{article}

\usepackage{acl}

\usepackage{times}
\usepackage{latexsym}
\usepackage{inconsolata}
\usepackage{amsfonts}
\usepackage{soul}
\usepackage{colortbl}
\usepackage{multirow}
\usepackage{xspace}
\usepackage{amsmath}
\usepackage{graphicx}
\usepackage{booktabs}

\newcommand{\FT}{\textsc{FineTune}\xspace}
\newcommand{\MASK}{\textsc{DebiasMask}\xspace}
\newcommand{\KW}{\textsc{KernelWhitening}\xspace}
\newcommand{\ETE}{\textsc{E2E-PoE}\xspace}

\newcommand{\IE}{\textsc{IEGDB}\xspace}
\newcommand{\READ}{\textsc{READ}\xspace}
\newcommand{\OursName}{\textsc{FairFlow}\xspace}
\newcommand{\OursPoe}{\OursName-\textsc{poe}\xspace}
\newcommand{\OursFocal}{\OursName-\textsc{focal}\xspace}
\newcommand{\OursCL}{\OursName}

\usepackage[T1]{fontenc}

\usepackage[utf8]{inputenc}

\usepackage{microtype}

\usepackage{inconsolata}

\title{\OursName: Mitigating Dataset Biases through Undecided Learning\\for Natural Language Understanding} 

\author{Jiali Cheng \and Hadi Amiri \\
  University of Massachusetts Lowell \\
  \texttt{\{jiali\_cheng, hadi\_amiri\}@uml.edu}
\\}

\begin{document}
\maketitle

\begin{abstract}
    
Language models are prone to dataset biases, known as shortcuts and spurious correlations in data, which often result in performance drop on new data. We present a new debiasing framework called ``\OursName'' that mitigates dataset biases by learning to be {\em undecided} in its predictions for data samples or representations associated with known or unknown biases. The framework introduces two key components: a suite of data and model perturbation operations that generate different biased views of input samples, and a contrastive objective that learns debiased and robust representations from the resulting biased views of samples. Experiments show that \OursName outperforms existing debiasing methods, particularly against out-of-domain and hard test samples without compromising the in-domain performance\footnote{Our code is available at \url{https://github.com/CLU-UML/FairFlow}.}.
\end{abstract}


\section{Introduction}

Existing computational models developed for natural language processing (NLP) tasks are vulnerable to dataset biases and spurious correlations in data, often referred to as ``shortcuts.''  
These shortcuts enable models to achieve high performance on NLP datasets by exploiting surface-level correlations between features and labels. However, they also result in a significant performance drop on hard or slightly modified test data~\citep{naik-etal-2018-stress}. For example, in the area of natural language inference (NLI), models like BERT~\citep{devlin-etal-2019-bert} tend to misclassify premise-hypothesis pairs that contain ``negation'' words in their hypotheses as ``contradiction,'' which happen to be predictive features associated with the \textit{contradiction} label in certain NLI datasets~\citep{gururangan-etal-2018-annotation,poliak-etal-2018-hypothesis,modarressi-etal-2023-guide}.


\begin{figure}[t]
    \centering
    \vspace{-20pt}
    \includegraphics[width=.45\textwidth]{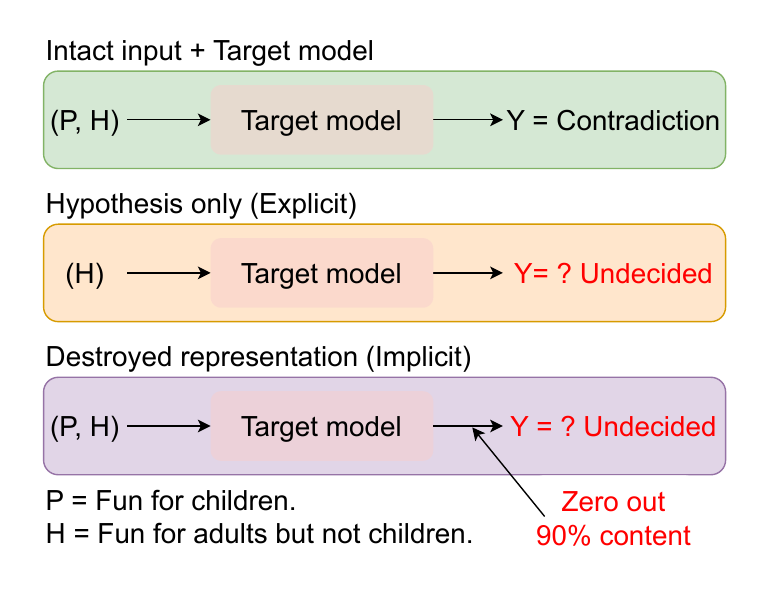}
    \caption{An example highlighting the concept of ``undecided learning'' using two types of data perturbation techniques. Given a premise-hypothesis pair in NLI, the model is expected to correctly classify their entailment relationship. However, given only the 
    hypothesis, a robust model should be undecided, i.e., refrain from making a definite judgment 
    about the relationship between an unknown premise and 
    the given hypothesis. Similarly, given a severely corrupted representation, a robust model should be undecided about the relation between a corrupted premise and hypothesis pair.
    Models that retain confidence in assigning labels to such inputs are likely to rely on shortcuts. \OursName takes an undecided stance against such inputs.}
    \label{fig:motivating_example}
    \vspace{-10pt}
    \label{fig:example}
\end{figure}

Existing debiasing approaches 
can detect known~\citep{clark-etal-2019-dont,sanh2020learning,karimi-mahabadi-etal-2020-end,modarressi-etal-2023-guide} and previously unidentified or unknown~\citep{utama-etal-2020-towards,sanh2020learning} biases in training data.
They mitigate dataset biases by 
re-weighting examples~\citep{sanh2020learning,karimi-mahabadi-etal-2020-end}, 
learning robust representations~\citep{gao-etal-2022-kernel,du-etal-2023-towards}, 
learning robust feature interaction patterns~\citep{wang-etal-2023-robust}, or 
reducing the effect of biased model components~\citep{meissner-etal-2022-debiasing}.


Despite the significant progress made in addressing dataset biases, existing models have certain limitations:
\textbf{(a)}: they often adopt a {\em single view} to dataset biases and primarily focus on specific types of biases~\citep{clark-etal-2019-dont,karimi-mahabadi-etal-2020-end}. However, rich sources and diverse types of dataset biases can be present in the data.
\textbf{(b)}: existing approaches that are based on weak learners~\citep{utama-etal-2020-towards,sanh2020learning,ghaddar-etal-2021-end,meissner-etal-2022-debiasing} rely on a {\em single} weak learner to identify biases, which inevitably tie their performance to the capabilities of the chosen weak learner.
\textbf{(c)}: prior works often evaluate debiasing methods using BERT-based models, which may limit their generalizability to other model architectures. 


We tackle the above challenges by developing \OursName--a multiview contrastive learning framework that mitigates dataset biases by being {\em undecided} in its prediction for biased views of data samples (see \textbf{Figure~\ref{fig:example}}). Specifically, the proposed method employs several data perturbation operators to generate biased views of intact data samples and integrate them into the training data and learning process. 
When presented with biased inputs, the model is trained to be undecided about the possible labels by making a uniform prediction across the label set. At the same time, the model is encouraged to be confident about intact inputs, which {\em often} serve as a reference for unbiased samples. Therefore, the approach encourages learning representations that are more attentive to the true signal of the underlying tasks rather than relying on shortcuts that are specific to certain datasets. 
In addition, the inherent randomness of the implicit perturbations in FairFlow (\S\ref{sec:imp_bias}) exposes the model to a diverse range of perturbations and prevents it from overfitting to specific types of biases present in the data.\looseness-1

The contributions of this paper are: 
\begin{itemize}
    \itemsep-1pt 
    \item categorization of dataset biases: we categorize prevalent data biases in NLU and model them using data perturbation operations;
    
    \item bias mitigation as an ``undecided learning'' problem: we formulate the bias mitigation problem as an ``undecided learning'' problem, which encourages reliance on genuine and task-related signals for effective debiasing;  
    \item robust performance on challengng samples: our approach shows robust results on {\em harder} test data while maintaining strong in-domain performance across several NLU tasks.\looseness-1
\end{itemize}

The experimental results show that \OursName obtains substantial improvement over competing models. Specifically, it achieves an average performance gain of 10.2 points on stress test datasets across several NLU tasks while maintaining performance on the original test sets. In addition, models trained using our framework show strong transferability, resulting in an average gain of 3.7 points in transfer testing experiments across different datasets and domains. 
Furthermore, 
we show that existing methods can be further improved by incorporating the proposed perturbation operators within their original objectives, resulting in a substantial average improvement of 5.8 points on stress test sets across datasets. 

\begin{figure*}[t]
    \centering
    \vspace{-20pt}
    \includegraphics[width=.9\textwidth]{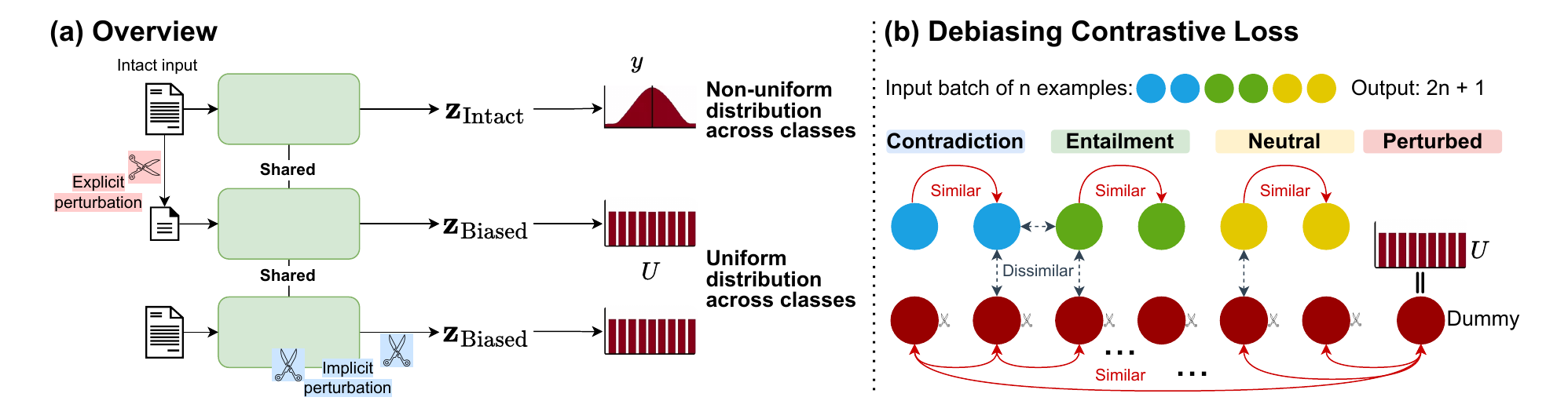}
    \caption{Architecture of the proposed model. (a) Explicit and implicit perturbations are applied to inputs to obtain biased prediction $z_{\mathrm{Biased}}$. (b) Biased predictions are drawn closer to uniform distribution, while predictions for intact input are pushed away from uniform distribution through contrastive learning.}
    \label{fig:model}
\end{figure*}

\section{Method}
\subsection{Problem Formulation}
We consider a dataset $\mathcal{D} = \{(x_i, y_i)|_{i=1}^n\}$, where $x_i$ is the $i$-th input consisting of several constituents $x_i = (x_i^1, x_i^2, \dots, x_i^p), |x_i| = p > 1$, and $y_i$ is the corresponding output for $x_i$. For example, in case of NLI, $p=2$ represents premise and hypothesis in each input and $y_i$ reflects the entailment or no-entailment relationship between the input pair. Our goal is to develop a model that is robust against different types of dataset biases in $\mathcal{D}$. 
We note that the model can be applied to a more general setting where input $x_i$ does not explicitly consist of several constituents, see \S\ref{sec:explicit_bias}.


\subsection{Overview}
We categorize dataset biases as \textit{explicit} and \textit{implicit} biases. Explicit biases are readily discernible and understandable by humans, such as high degree of lexical overlap between the premise and hypothesis in case of NLI. On the other hand, implicit biases are often subtle, indiscernible to humans, and more challenging  
to detect. For example, any word in input has the potential to act a shortcut, resulting in spurious correlations. 
We introduce different types of explicit and implicit biases that are {\em task-independent} and generally applicable to bias mitigation across NLP datasets (\S\ref{sec:biasmodeling}). 
Given such categorization, we propose a debiasing framework that mitigates dataset biases by learning genuine task-related representations that are attentive to the true signal of the tasks rather than biases and shortcut solutions in datasets. 
The key novelty of our approach is in imposing a downstream model to adopt an ``undecided'' (``uncertain'') stance in its predictions when presented with biased views of inputs. The framework achieves this goal by assigning a uniform probability across the labels, see \textbf{Figure~\ref{fig:model}}. 
Specifically, the model regularizes the loss of the target task with a contrastive loss which draws biased predictions closer to a uniform distribution while pushing other predictions away from uniform distribution (\S\ref{sec:contrastive}).  
%





\subsection{Bias Modeling}\label{sec:biasmodeling}
We present a series of data perturbation operations to generate biased views by corrupting intact inputs. These perturbations can be explicit or implicit. In explicit perturbation, we directly corrupt the input data, while in implicit perturbation, we corrupt the representations of the input data. These perturbation techniques impose controlled variations on the data, which enable us to conduct a thorough analysis of their effects on bias mitigation.


\subsubsection{Explicit Biases} \label{sec:explicit_bias}


\paragraph{Ungrammatical Perturbation} Recently, \citet{sinha-etal-2021-unnatural} showed that traditional and recent neural language models can be largely invariant to random word order permutation in their inputs. 
An ungrammatical input is often not understandable by humans and can potentially lead to explicit biases when models confidently predict outcomes for such inputs. For example, a model making a confident prediction about the contradiction class for the following perturbed premise-hypothesis pair from Figure~\ref{fig:example} may attribute its confidence to the negation term in the hypothesis: (``{\tt children fun for}'', ``{\tt children fun adults but for not}''). 
To obtain an input with grammatical biases, we design the perturbation operation $\mathcal{P}_{Gra}$ that 
corrupts the word order in each input $x_i$. We encode the shuffled input using the shared encoder $f$ and transform it with a branch-specific MLP as follows:
\begin{equation}
    z_{Gra} = \texttt{MLP}_{Gra} \Big( f\big(\mathcal{P}_{Gra}(x_i) \big) \Big).
\end{equation}

\paragraph{Sub-input Perturbation} 
In NLP tasks that involve multi-part inputs (such as NLI), it is crucial to use the information from all parts of the input for prediction, i.e., all constituents should collectively contribute to accurate prediction. More importantly, an incomplete input should not lead to a confident prediction, as important information may be removed. 
Therefore, an explicit bias arises when the model makes confident predictions based on incomplete input, such as predicting the \textit{entailment} relation when only the hypothesis is provided as input in case of NLI. Sub-input biases can arise from any part of the input, denoted as $\{x_i^j\}_{j=1}^p$, or from various text spans within different sub-parts.
To realize sub-input biases, we define the $\mathcal{P}_{Sub}$ operator that takes one of the constituents of $x_i$, which is hen encoded with a shared encoder $f$ and further transferred with a constituent-specific $\texttt{MLP}_{Sub}$ as follows:
\begin{equation}
\label{eq:sub_aug}
    z_{Sub} = \texttt{MLP}_{Sub}\Big( f\big (\mathcal{P}_{Sub}(x_i)\big) \Big).
\end{equation}

\noindent We note that this operator is applicable to a more general setting where input $x_i$ does not explicitly consist of several constituents, e.g., in general text classification problems. In such cases, each $x_i$ can be divided into $p>1$ text segments. However, we acknowledge that there are tasks in which one sub-input, i.e. ${x_i^j}$ for a specific $j$, is enough to make a correct prediction for the complete input ${x_i}$, and therefore remaining undecided may seem counter-intuitive. Nevertheless, by training the model to be undecided when presented with incomplete information, we minimize the risk of biased predictions based solely on partial information, which can, in turn, make the model more robust against potential biases associated with incomplete data.

\subsubsection{Implicit Biases}
\label{sec:imp_bias}
The idea of implicit perturbations is to obtain biased representations of intact data, without explicitly perturbing the input. We introduce model- and representation-based implicit perturbation.

\paragraph{Model-based Perturbation} This approach largely perturbs a given model by converting it into a much weaker model, using mechanisms such as sparsification and layer dropping~\citep{NEURIPS2021_6e8404c3}. A weaker model is believed to capture more biases than a stronger model~\citep{ghaddar-etal-2021-end,sanh2020learning,utama-etal-2020-towards}. While existing methods require training a weak learner in advance~\citep{utama-etal-2020-towards,sanh2020learning,meissner-etal-2022-debiasing}, our method obtains biased predictions through the same deep neural model ($f$) and can be trained \emph{end-to-end}. Formally, we design a model-based perturbation operator $\mathcal{P}_{Mod}$ that uses only the first $k$ layers of the shared encoder $f$, which results in a substantially weakened model with reduced representation power. This branch encodes the intact input using the perturbed model and transform it with a branch-specific MLP as follows:
\begin{equation}
    z_{Mod} = \texttt{MLP}_{Mod} \Big( \mathcal{P}_{Mod}(f)(x_i) \Big).
\end{equation}

\paragraph{Representation-based Perturbation} This perturbation encodes the intact input with the original encoder $f$ but significantly corrupts the generated representations. Given this severely damaged and much less meaningful representation, the model should not be able to predict the correct label. We design a representation-based perturbation operator $\mathcal{P}_{Rep}$ that corrupts the intact representation, $f(x_i)$, and creates a severely perturbed representation. We then transform the perturbed representation with a branch-specific MLP as follows:
\begin{equation}
\label{eq:rep_aug}
    z_{Rep} = \texttt{MLP}_{Rep} \Big( \mathcal{P}_{Rep} \big( f(x_i) \big) \Big).
\end{equation}


\textbf{Table~\ref{tab:aug_list}} summarizes the above perturbation operators and provides details of their implementations. 

\subsection{Supervised Contrastive Debiasing}\label{sec:contrastive}
Given the explicit and implicit biased views of data samples, we expect a robust debiasing model to maintain an ``undecided'' stance across labels for biased inputs while providing confident predictions for intact inputs $x_i, \forall i$. Based on this intuition, the outputs of the bias branches should approximate a {\em uniform distribution} ($U$) across classes, while the output of the original branch should align with its corresponding gold distribution, i.e., the label $y_i$. 
To achieve this goal, we adapt the supervised contrastive loss~\citep{khosla2020supervised}, which operates by first grouping samples based on their respective labels, and then encouraging predictions (logits) of pairs within the same group to become closer while pushing logits of pairs across different groups further apart, i.e. forming positive pairs within the same group while creating negative pairs using all other pairs: 

\begin{table}\small
\centering
\setlength{\tabcolsep}{3pt}
\renewcommand{\arraystretch}{1} 
\begin{tabular}{c|c|l}
\toprule
\textbf{Operator}               & \textbf{Type}     & \textbf{Implementation} \\
\midrule
$\mathcal{P}_{Gra}$ & Explicit & Shuffle tokens in $x_i$ randomly \\
$\mathcal{P}_{Sub}$  & Explicit & Drop $1/p$ of tokens from $x_i^j$ randomly \\
$\mathcal{P}_{Sub}$  & Explicit & Drop $x_i^j, j=1\dots p$ \\
\midrule
$\mathcal{P}_{Mod}$  & Implicit & Use only first $k$ of layers of $f$\\
$\mathcal{P}_{Rep}$  & Implicit & Zero out $m\%$ of values in $f(x_i)$ \\
\bottomrule
  \end{tabular}
  \caption{Implementations of proposed perturbations}
  \label{tab:aug_list}
  \vspace{-10pt}
\end{table}

We adapt this loss function for bias mitigation as follows (described for a single perturbation for simplicity):
given a batch of $n$ non-perturbed examples, we perturb them using a perturbation technique described in Table~\ref{tab:aug_list}. The perturbed examples form a single group as they all have the same label (a uniform distribution across all classes), and the non-perturbed examples with the same label form separate groups.\footnote{For example, four groups in case of NLI: perturbed examples, non-perturbed examples labeled as `entailment', non-perturbed examples labeled as `contradiction', and non-perturbed examples labeled as `neutral'.} 
As illustrated in \textbf{Figure~\ref{fig:model}}, we encourage the model to be undecided about the label of perturbed inputs by adding a dummy example that has a ``fixed'' uniform distribution across all labels to the group of perturbed examples, resulting in a batch of $2n+1$ examples ($\mathcal{I}$). We compute the contrastive loss as follows:
\begin{multline}
\label{eq:supcontras}
    \mathcal{L}_{\mathrm{Debias}} = \\ 
    \sum_{i \in \mathcal{I}} \frac{-1}{|\mathcal{G}(i)|} \sum_{j \in \mathcal{G}(i)} \mathrm{log} \frac{\mathrm{exp} (z_i \cdot z_j / \tau)}{\sum_{k \in \mathcal{A}(i)} \mathrm{exp} (z_i \cdot z_k / \tau)},
\end{multline}
where $\mathcal{G}(i)$ is the set of examples that are in the same group as $i$ (having the same label as $i$); 
$\mathcal{A}(i) = \mathcal{I}\backslash\{i\}$ is the set of all examples except $i$; 
$z$ indicates the logit of an example, which for perturbed examples is obtained from one of the Equations (\ref{eq:sub_aug})--(\ref{eq:rep_aug}); and 
$\tau$ denotes the temperature parameter.\footnote{We note that the summation over all samples except $i$ in the denominator of (\ref{eq:supcontras}) is motivated by noise contrastive estimation and N-pair losses~\citep{khosla2020supervised,pmlr-v9-gutmann10a,NIPS2016_6b180037}, in which the ability to discriminate between signal and noise (negative class) is improved by adding more examples of negative class.} The dummy example in the perturbed group has a fixed uniform distribution across all labels as its $z$. This formulation encourages the model to be undecided about the label of perturbed inputs, while being confident about the labels of intact inputs, allowing it to effectively distinguish between different groups of examples. 

Finally the model learns the debiasing task in an \emph{end-to-end} manner by minimizing the standard cross-entropy loss with predictions of intact input $z_{\mathrm{Intact}}=f(x_i)$ and the debiasing loss, weighted by a balancing hyperparameter $\lambda$ as follows:
\begin{equation}
    \theta^{*} = \arg\min_{\theta} \mathcal{L}_{\mathrm{CE}}(z_{\mathrm{Intact}}, y_i) + \lambda \mathcal{L}_{\mathrm{Debias}}.
\end{equation}

\paragraph{Compatibility and Difference with Other Debiasing Objectives and Training Methods} Our framework is designed to be 
compatible with debiasing objectives in existing literature. Notably, it can incorporate objectives such as the product of experts (PoE)~\citep{karimi-mahabadi-etal-2020-end,clark-etal-2019-dont}, debiased focal loss~\citep{karimi-mahabadi-etal-2020-end}, and other possible objectives, see Appendix~\ref{sec:debias_obj} for more details. In experiments, we show that our framework can further improve these well-performing baseline models. One major difference with existing debiasing objectives is that prior works use a biased model to measure how much biases present in input, while \OursCL encourages robust models to be undecided given known biased inputs, obtained by the proposed perturbations.
Moreover, we do not impose any restriction on the parametrization of the underlying model $f$, making our framework flexible to work with a wide range of training methods and network architectures (Table~\ref{tab:roberta}-\ref{tab:gpt2} in Appendix).

\section{Experiments}

\paragraph{Setup}
We employ BERT~\citep{devlin-etal-2019-bert} as the commonly-used base model in previous works. In addition, we extend our evaluation to RoBERTa~\citep{liu2019roberta} and GPT-2~\citep{radford2019language} for a more comprehensive analysis.

\paragraph{Datasets} We evaluate our debiasing framework on three NLP datasets including 
MNLI~\citep{williams-etal-2018-broad}, 
paraphrase identification using Quora question pairs (QQP)~\citep{sharma2019natural}, and
relation extraction using gene-phenotype relation (PGR)~\citep{sousa-etal-2019-silver}. 
These datasets are used for {\em in-domain} (ID) evaluation.

\paragraph{Stress Test Sets} We assess the robustness of models against spurious correlations using ``stress test sets,'' specifically designed with hard examples to challenge models. 
We use the stress test set for MNLI from~\citep{naik-etal-2018-stress}, and
use the same approach to generate the stress test set for QQP.
For PGR, the label-preserving rules from previous tasks do not apply due to the nature of this dataset.
However, given the long-tail distribution of entity appearances, we create a stress test set for PGR by selecting test examples in which both entities appear less than five times in the training set.

\paragraph{OOD Test Sets} We assess the performance of models on existing out-of-distribution (OOD) test sets, which serve as another challenge benchmark. For MNLI, we use HANS~\citep{mccoy-etal-2019-right}, which is designed to test models' capabilities against lexical and syntactic heuristics in data. For QQP, we employ the PAWS dataset~\citep{zhang-etal-2019-paws}, which focuses on paraphrase identification in cases of high lexical and surface-level similarity between question pairs.

\paragraph{Transfer Test Sets} We evaluate the performance of models in maintaining strong transferability across datasets. We use SNLI~\citep{bowman-etal-2015-large} and MRPC~\citep{dolan-brockett-2005-automatically} as the transfer set for MNLI and QQP, respectively. 

\begin{table*}
\tiny
\centering
\begin{tabular}{l|cccc|cccc|cc|cccc}
\toprule
\multirow{2}{2em}{\textbf{Model}} & \multicolumn{4}{c|}{\textbf{MNLI} (Acc.)} & \multicolumn{4}{c|}{\textbf{QQP} (F1)} & \multicolumn{2}{c|}{\textbf{PGR} (F1)} & \multicolumn{4}{c}{\textbf{Avg.}}\\
               & ID   & Stress & OOD & Transfer & ID & Stress & OOD & Transfer & ID & Stress & ID & Stress & OOD & Transfer \\ 
    \toprule
    \textbf{\FT}        & 84.3 & 61.7 & 59.7 & 78.7 & 88.6 & 63.3 & 47.7 & 65.1 & 64.3 & 55.2 & 79.1 & 60.1 & 53.7 & 71.9 \\ 
    \midrule
    \textbf{\MASK}      & 83.5 & 59.7 & 59.7 & 78.3 & 88.1 & 64.6 & 50.3 & 68.5 & 64.1 & 51.7 & 78.6 & 58.7 & 55.0 & 73.4 \\ 
    \textbf{\KW}        & 84.0 & 60.9 & 60.2 & 78.4 & 88.8 & 65.1 & 51.2 & 69.6 & 64.3 & 51.8 & 79.0 & 59.3 & 55.7 & 74.0 \\ 
    \textbf{\ETE}       & 83.4 & 61.3 & 62.3 & 77.5 & 88.5 & 64.5 & 51.4 & 70.5 & 63.0 & 53.6 & 78.3 & 59.8 & 56.8 & 74.0 \\
    \textbf{LSWC}       & 80.7 & 59.4 & 59.3 & 77.7 & 87.1 & 65.8 & 49.6 & 70.0 & 63.3 & 52.8 & 77.0 & 59.3 & 54.5 & 73.8 \\ 
    \textbf{\IE}        & 84.1 & 61.8 & 62.7 & 78.1 & 87.6 & 63.5 & 53.0 & 68.3 & 64.2 & 54.9 & 78.6 & 60.1 & 57.9 & 73.2 \\
    \textbf{\READ}      & 80.8 & 61.5 & 63.4 & 75.1 & 87.0 & 66.7 & 53.6 & 68.2 & 63.0 & 54.4 & 76.9 & 60.9 & 58.5 & 71.7 \\ 
    \midrule
    \textbf{\OursPoe}   & 84.6 & 64.3 & 64.3 & 79.5 & 88.8 & 71.0 & 53.9 & 70.4 & 64.9 & 55.9 & 79.4 & 63.7 & 59.1 & \underline{75.0} \\ 
    \textbf{\OursFocal} & \underline{84.9} & \underline{64.8} & \underline{64.3} & \underline{79.3} & \underline{89.5} & \underline{71.3} & \underline{54.9} & \underline{70.7} & \underline{65.4} & \underline{56.5} & \underline{79.9} & \underline{64.2} & \underline{59.6} & \underline{75.0} \\
    \textbf{\OursCL}    & \textbf{85.1} & \textbf{65.4} & \textbf{64.9} & \textbf{79.6} & \textbf{90.4} & \textbf{72.0} & \textbf{56.0} & \textbf{72.4} & \textbf{65.9} & \textbf{56.6} & \textbf{80.5} & \textbf{64.7} & \textbf{60.5} & \textbf{76.0} \\ 
    \bottomrule
  \end{tabular}
  \caption{Experimental results on three datasets averaged across three architectures. Results for each architecture are shown in Table~\ref{tab:bert}-\ref{tab:gpt2} in Appendix. The best performance is in \textbf{bold} and the second best is \underline{underlined}.}
  \label{tab:main}
\end{table*}

\begin{table*}
\small
\centering
\begin{tabular}{l|cccc}
\toprule
\multirow{2}{*}{\textbf{Model}} & \multicolumn{4}{c}{\textbf{Avg.}}\\
                & ID & Stress & OOD & Transfer \\ 
    \midrule
    \textbf{\FT}        & 79.1 & 60.1 & 53.7 & 71.9 \\ 
    \midrule
    \textbf{\ETE}       & 78.3 & 59.8 & 56.8 & 74.0 \\
    \textbf{\MASK}      & 78.6 & 58.7 & 55.0 & 73.4 \\ 
    \textbf{LSWC}       & 77.0 & 59.3 & 54.5 & 73.8 \\ 
    \textbf{\IE}        & 78.6 & 60.1 & 57.9 & 73.2 \\
    \textbf{\KW}        & 79.0 & 59.3 & 55.7 & 74.0 \\ 
    \textbf{\READ}      & 76.9 & 60.9 & 58.5 & 71.7 \\ 
    \midrule
    \textbf{\OursPoe}   & 79.4 & 63.7 & 59.1 & \underline{75.0} \\ 
    \textbf{\OursFocal} & \underline{79.9} & \underline{64.2} & \underline{59.6} & \underline{75.0} \\
    \textbf{\OursCL}    & \textbf{80.5} & \textbf{64.7} & \textbf{60.5} & \textbf{76.0} \\ 
    \bottomrule
  \end{tabular}
\end{table*}

\paragraph{Baselines} 
We consider the following baselines:
\vspace{-5pt}
\begin{itemize}
    \itemsep-2pt
    \itemindent-10pt
    \item \textbf{\FT} standard finetuning without debiasing based on the base model used. 
    \item \textbf{\ETE}~\citep{karimi-mahabadi-etal-2020-end}, which trains a biased model on the hypothesis only and trains a robust model using Product of Experts (PoE)~\citep{10.1162/089976602760128018}.
    \item \textbf{\MASK}~\citep{meissner-etal-2022-debiasing}, which first trains a weak learner and then prunes the robust model using PoE. 
    \item \textbf{\KW}~\citep{gao-etal-2022-kernel}, which learns isotropic sentence embeddings using Nystr\"{o}m kernel approximation~\citep{Xu_Jin_Shen_Zhu_2015} method, achieving disentangled correlation between robust and spurious embeddings.
    \item \textbf{LWBC}~\citep{kim2022learning}, which learns a debiased model from a commitee of biased model obtained from subsets of data.
    \item \textbf{\IE}~\citep{du-etal-2023-towards}, which mitigates dataset biases with an ensemble of random biased induction forest; the model induces a set of biased features and then purifies the biased features using information entropy\footnote{While this method does not have a publicly released code, we tried our best to reproduce their approach and results with a few points lower than reported.}.
    \item \textbf{\READ}~\citep{wang-etal-2023-robust}, which assumes that spuriousness comes from the attention
    and proposes to do deep ensemble of main and biased model at the attention level to learn robust feature interaction.
\end{itemize}

\section{Results and Discussions}

\paragraph{Robust Debiasing Model}
The main results in \textbf{Table~\ref{tab:main}} shows our model with three objectives: contrastive learning (\OursCL), product of experts (\OursPoe) and focal loss (\OursFocal), see \S\ref{sec:contrastive}. They all achieve high performance across all datasets and test sets including in-domain (ID), stress, and out-of-distribution (OOD) test sets. By adopting the undecided learning objective, the model learns debiased and robust representations without loss of in-domain performance. 
Across three datasets, our best-performing model (\OursCL) outperforms \MASK, \KW, \ETE, \IE, \READ approaches by 2.0, 6.1 and 5.5; 1.5, 5.5 and 4.7; 2.2, 4.9 and 3.6; 1.9, 4.7 and 2.7; 3.6, 3.8 and 1.9 absolute points on the ID, stress and OOD test sets respectively. We attribute these gains to the use of biased branches and undecided learning, realized through the proposed contrastive objective. 

We note that \IE and \READ provide debiasing gains at the cost of ID performance, with a performance drop of 0.2, 1.0 and 0.1; 3.5, 0.4 and 1.3 compared to \FT on MNLI, QQP, PGR respectively. Specifically, we attribute the large performance drop of \READ to the deep ensemble (compared to logit ensemble of \ETE and \OursPoe) of the target and biased model at the attention level, which may impose excessive regularization on the model. However, our model learns robust representations without loss on ID test sets across all three objectives. 

In addition to better debiasing performance, our approach shows stronger transferability compared to baselines. Specifically, \OursCL outperforms \MASK, \KW, and \ETE on transfer test set by 2.7, 2.1 and 2.1, respectively. In addition, \OursPoe and \OursFocal retain strong transfer performance as well, indicating that our framework does not hurt models' transferability.

Comparing different fusion techniques in the last three rows in Table~\ref{tab:main}, we observe that the proposed contrastive objective is more effective than PoE~\citep{karimi-mahabadi-etal-2020-end,clark-etal-2019-dont,sanh2020learning} and debias focal loss~\citep{karimi-mahabadi-etal-2020-end}, in particular on stress and OOD test sets. We also find that debias focal loss almost always outperform PoE on our datasets, which is inline with previous report by~\citet{karimi-mahabadi-etal-2020-end}.
%

\paragraph{More Bias Branches, Less Biased Model}
Unlike existing approaches that have a single view to dataset biases, our model employs multiple views, allowing it to effectively capture and mitigate various types of biases present in the data. 
%
Specifically, compared to \ETE which only captures one sub-input bias, \OursPoe achieves on average 1.8, 9.5 and 3.8 absolute points improvement on ID, stress and OOD test set across three different datasets. Both methods employ PoE as the fusion technique. Compared to \MASK~\citep{meissner-etal-2022-debiasing} which only captures bias though a weak model, \OursPoe achieves 1.5, 12.3 and 11.0 points improvement on ID, stress and OOD test sets, respectively.

\begin{table}
\small
\centering
\begin{tabular}{l|cccc}
    \toprule
    Model & ID & Stress & OOD & Transfer \\ 
    \midrule
    \textbf{No debiasing}      & 84.6 & 57.3 & 56.2 & 80.3 \\
    \midrule
     + DropPremise    & 84.6 & 61.6 & 65.5 & 80.6 \\
     + DropHypothesis & 84.6 & 61.6 & 66.3 & 80.6 \\
     + HalfHalf       & 84.8 & 62.1 & 64.2 & 80.0 \\
     + Shuffle        & 84.8 & 62.1 & 63.9 & 80.0 \\
    \midrule
     + DropLayer      & 84.8 & 62.0 & 65.4 & 80.4 \\
     + DestroyRep     & 84.8 & 62.3 & 66.5 & 80.0 \\
    \midrule
    \midrule
    \textbf{Full model}        & 84.9 & 63.6 & 68.4 & 81.1  \\
    \midrule
     - DropPremise    & 84.6 & 61.6 & 63.2 & 80.6 \\
     - DropHypothesis & 84.6 & 61.6 & 62.6 & 80.6 \\
     - HalfHalf       & 84.8 & 62.1 & 63.8 & 80.0 \\
     - Shuffle        & 84.8 & 62.1 & 65.3 & 80.0 \\
    \midrule
     - DropLayer      & 84.5 & 60.5 & 62.5 & 80.4 \\
     - DestroyRep     & 84.5 & 60.5 & 62.7 & 80.4 \\
    \bottomrule
\end{tabular}
\caption{Contribution of each perturbation branch in our method on MNLI.}
\label{tab:ablation}
\end{table}

\paragraph{Branches Contribute Differently}
To examine the contribution of each perturbation branch, we conduct ablation studies on MNLI. Specifically, we add one branch at a time to the vanilla model or remove one branch at a time from the full model, see \textbf{Table~\ref{tab:ablation}}. The perturbations include DropPremise and DropHypothesis, which drop the premise and hypothesis from the input respectively; HalfHalf, which randomly drops $k=50\%$ of the tokens from input; Shuffle, which randomly shuffles the input; 
DropLayer, which drops all layers after the 2nd layer; and
DestroyRep, which zeros out $m=90\%$ of the elements in the intact representation. 
The results show that all perturbations contribute positively to the overall performance on ID, stress, OOD, and transfer test sets. Specifically, explicit perturbations can improve the vanilla model on average by 0.1 and 4.6 absolute point on ID and stress test sets respectively. While implicit perturbations improve the vanilla model on average by 0.1 and 4.9 points. In addition, DestroyRep achieves the best performance on the  stress and OOD test sets, while DropPremise and DropHypothesis achieve the best performance on the transfer set. 

\begin{figure}[t]
    \centering
    \vspace{-10pt}
    \includegraphics[width=0.5\textwidth]{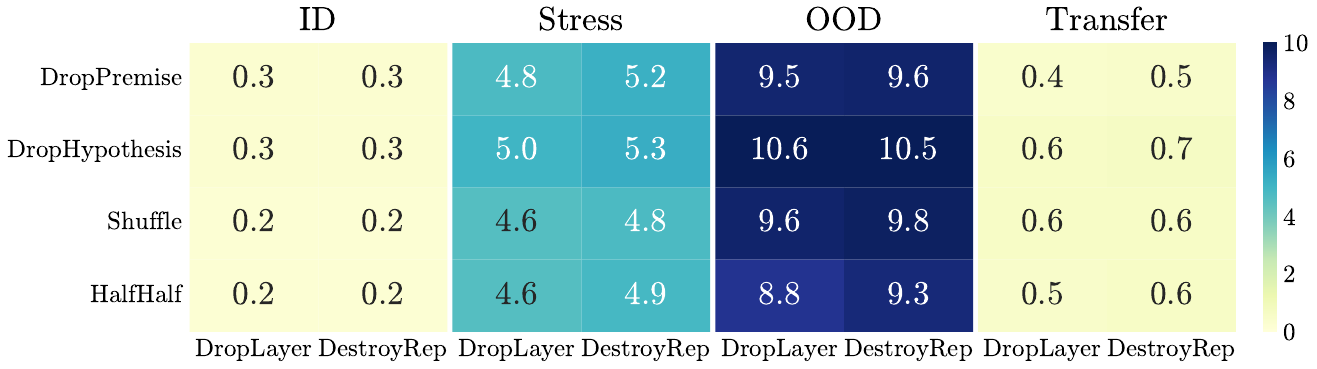}
    \vspace{-20pt}    
    \caption{Debiasing performance with different combinations of explicit and implicit perturbations. 
    The values indicate relative accuracy 
    increase compared to vanilla fine-tuning.}
    \label{fig:two_aug}
\end{figure}

In addition, we investigate the effect of different combinations of perturbations. Specifically, we train our model with one explicit perturbation and one implicit perturbation at a time. \textbf{Figure~\ref{fig:two_aug}} illustrates the relative increase of performance to standard fine-tuning across ID, stress and OOD test sets. 
Two combinations yields better results on the OOD test set. The first combines DropPremise or DropHypothesis with DropLayer, while the second combines perturbation of all inputs (e.g. Shuffle) and PurturbRep. The improved results likely stem from the complementary strengths of these diverse perturbation techniques, which can create a more robust debiasing model. 

\paragraph{Debiased Models Are Still Biased}
Our results in Table~\ref{tab:main} and prior reports~\citep{mendelson-belinkov-2021-debiasing,ravichander-etal-2023-bias} show that debiased methods can still be biased. For example, \MASK and \KW show higher levels of biases than \FT
by 3.7 and 4.2 points on the stress test set~\citep{naik-etal-2018-stress} respectively. These results emphasize the need for modeling multiple types of biases, and highlights the advantages of our approach.

\paragraph{\OursCL Maintains Generalization across Biases}
Bias in existing methods may be because of their tendency to over-specialize in specific types of biases. \textbf{Table~\ref{tab:subset}} summarizes the performance of debiasing models across different subsets of the stress set. \OursCL achieves the maximum average performance with smaller standard deviation across these subsets, indicating that it does not overfit to specific biases. We attribute such resilience to \OursCL's incorporation of both explicit and implicit perturbations, along with the randomness in implicit perturbations, which allows the model to effectively handle diverse set of biases.





\begin{table}
\footnotesize
\centering
\begin{tabular}{l|lc}
\toprule
\textbf{Model} & \textbf{Param} & \textbf{Time (hr)} \\
\midrule
    \textbf{\FT}   & 110M + 2K & 4.2 \\
    \midrule
    \textbf{\MASK} & + 28M + 2K & 5.3 \\
    \textbf{\KW}   & + 3K & 6.3 \\
    \textbf{\ETE}  & + 30K & 5.5 \\
    \textbf{\IE}   & + 50 $\times$ 2K & 7.2 \\
    \textbf{\READ} & + 28M + 2K & 4.9 \\
    \midrule
    \textbf{\OursCL} & + 2 $\times$ 2K & 4.9 \\
    \bottomrule
  \end{tabular}
  \caption{Efficiency of debiasing models on MNLI.}
  \vspace{-20pt}
  \label{tab:eff}
\end{table}

\paragraph{Efficiency}
We evaluate the efficiency of different debiasing methods in terms of number of trainable parameters and training time. 
As \textbf{Table~\ref{tab:eff}} shows,  \OursCL introduces only 4K additional parameters, which is significantly less than 100K in \IE with 50 classifiers, and 28M in \MASK and \READ with an extra weak model. This highlights the efficiency gains from the proposed perturbation operations. Furthermore, \OursCL has the shortest training time. \OursCL achieves these efficiencies without requiring additional training data, operating only by generating diverse views of the input data.

\paragraph{Perturbation for Data Augmentation}
The explicit perturbation operators proposed in our framework offer a valuable opportunity for data augmentation, leading to improved performances on existing debiasing methods (See Table~\ref{tab:res_aug} in Appendix).

\paragraph{Bias in Different Parts of Inputs}
In our experiments with single explicit perturbations, we find that DropPremise and DropHypothesis lead to similar performances on MNLI, 
showing that there exists dataset bias in premise, potentially as much as those in hypothesis. However, many existing methods tend to overlook biases in the premise in NLI datasets.
In addition, biases can often emerge from the interplay of various parts of inputs, rather than a single source. HalfHalf and Shuffle perturbations can capture such types of biases by perturbing the entire inputs. We note that while additional weak learners can potentially capture biases from multiple sources~\citep{utama-etal-2020-towards,sanh2020learning,meissner-etal-2022-debiasing}, their effectiveness  is likely limited by the capabilities of the weak models. 
Our approach addresses dataset biases through a multiview approach to bias, which leads to a more robust debiasing process.

\section{Related Work}

\paragraph{Quantifying Bias}
Several works focus on understanding dataset bias and deibasing algorithms, including measurement of bias of specific words with statistical test~\citep{gardner-etal-2021-competency}, identification of biased and generation of non-biased samples with $z$-filtering~\citep{wu-etal-2022-generating}, identification of bias-encoding parameters~\citep{yu-etal-2023-unlearning}, when bias mitigation makes model less or more biased~\citep{ravichander-etal-2023-bias}, bias transfer from  other models~\citep{jin-etal-2021-transferability}, and  representation fairness~\citep{shen-etal-2022-representational}.

\paragraph{Debiasing with Biased Models} These approaches model shortcuts from datasets, and use biased predictions as a reference to quantify bias in input data. Bias can be \emph{explicit bias} in NLI datasets~\citep{belinkov-etal-2019-dont,clark-etal-2019-dont,karimi-mahabadi-etal-2020-end,utama-etal-2020-mind}, and \emph{implicit bias} detected by weak models~\citep{ghaddar-etal-2021-end,sanh2020learning,meissner-etal-2022-debiasing,utama-etal-2020-towards,meissner-etal-2022-debiasing}. 
Ensemble techniques include Product-of-Experts (PoE)~\citep{10.1162/089976602760128018,sanh2020learning,cheng24c_interspeech} which takes element-wise multiplication of the logits, Debiased Focal Loss~\citep{karimi-mahabadi-etal-2020-end} and ConfReg~\citep{utama-etal-2020-mind} which both down-weight predictions based on the confidence of biased models.\looseness-1

\paragraph{Debiased Representations} Existing methods focus on weak-learner guided pruning~\citep{meissner-etal-2022-debiasing}, disentangling robust and spurious representations~\citep{gao-etal-2022-kernel}, 
decision boundaries~\citep{lyu2023feature}, and
attention patterns with PoE~\citep{wang-etal-2023-robust}, 
training biased models with one-vs-rest approach~\citep{jeon-etal-2023-improving}, and amplifying bias in training set with debiased test set~\citep{reif-schwartz-2023-fighting}.


\paragraph{Fairness and Toxicity} These approaches focus on protected variable such as race. Existing methods spans across counterfactual data augmentation~\citep{zmigrod-etal-2019-counterfactual,dinan-etal-2020-queens,barikeri-etal-2021-redditbias}, comparisons between network architectures~\citep{meade-etal-2022-empirical}, deibasing with counterfactual inference~\citep{qian-etal-2021-counterfactual}, adversarial training~\citep{madanagopal-caverlee-2023-bias}, prompt perturbation~\citep{guo-etal-2023-debias}, data balancing~\citep{han-etal-2022-balancing}, contrastive learning~\citep{cheng2021fairfil}, detecting toxic outputs~\citep{schick-etal-2021-self}, performance degradation incurred by debiasing methods~\citep{meade-etal-2022-empirical}, and benchmarks~\cite{nadeem-etal-2021-stereoset,hartvigsen-etal-2022-toxigen,sun-etal-2022-chapterbreak}. Social debiasing methods may underperform in OOD settings because OOD examples may not contain social stereotypes or biases.



\section{Conclusion}
We investigate bias mitigation in NLU datasets by 
formulating the debiasing problem within a contrastive learning framework, incorporating explicit and implicit perturbation techniques and introducing undecided learning. Through extensive experiments across a range of NLU tasks, we demonstrate the effectiveness of our method in achieving improved debiasing performance, while maintaining performance on in-domain test sets. We find that existing methods (including ours) are still sensitive to dataset biases, and our experiments show the limitations of these approaches in fully addressing dataset biases. These results necessitate investigating a more systematic evaluation benchmark for debiasing. 
Our approach can potentially be improved by
investigating more complex biases~\citep{yao2023large,gandikota2023erasing}, 
exploring alternative training paradigms such as curriculum learning~\citep{bengio2009curriculum,vakil-amiri-2022-generic}, and evaluating robustness to unseen biases~\citep{NEURIPS2023_b0d9ceb3}. 
Beyond NLU, our work can potentially be applied to a broader range of applications~\citep{cheng2024mu,Liu_2024_WACV}.

\newpage
\section*{Limitations}
Though our framework outperforms baselines, there is still room for improvement on Stress and OOD test sets.   
In addition, we didn't analyze the generalizability of the approach to other NLP domains or tasks beyond the three tasks used in the experiments.  
\section*{Ethic and Broader Impact Statements}
Our research focuses on mitigating dataset biases in NLP datasets. There are no specific ethical concerns directly associated with this work. However, we acknowledge and emphasize the ethical mindfulness throughout the design, training, and applying the models investigated in this study on any applications.
The broader impacts of our work are in advancing dataset fairness and potentially enhancing decision-making based on data. By addressing biases, we contribute to improving the reliability of NLP datasets and the accuracy and transferability of the models trained using NLP datasets. 



\bibliography{anthology1,anthology2,custom}

\begin{thebibliography}{65}
\expandafter\ifx\csname natexlab\endcsname\relax\def\natexlab#1{#1}\fi

\bibitem[{Barikeri et~al.(2021)Barikeri, Lauscher, Vuli{\'c}, and
  Glava{\v{s}}}]{barikeri-etal-2021-redditbias}
Soumya Barikeri, Anne Lauscher, Ivan Vuli{\'c}, and Goran Glava{\v{s}}. 2021.
\newblock \href {https://doi.org/10.18653/v1/2021.acl-long.151}
  {{R}eddit{B}ias: A real-world resource for bias evaluation and debiasing of
  conversational language models}.
\newblock In \emph{Proceedings of the 59th Annual Meeting of the Association
  for Computational Linguistics and the 11th International Joint Conference on
  Natural Language Processing (Volume 1: Long Papers)}, pages 1941--1955,
  Online. Association for Computational Linguistics.

\bibitem[{Belinkov et~al.(2019)Belinkov, Poliak, Shieber, Van~Durme, and
  Rush}]{belinkov-etal-2019-dont}
Yonatan Belinkov, Adam Poliak, Stuart Shieber, Benjamin Van~Durme, and
  Alexander Rush. 2019.
\newblock \href {https://doi.org/10.18653/v1/P19-1084} {Don{'}t take the
  premise for granted: Mitigating artifacts in natural language inference}.
\newblock In \emph{Proceedings of the 57th Annual Meeting of the Association
  for Computational Linguistics}, pages 877--891, Florence, Italy. Association
  for Computational Linguistics.

\bibitem[{Bengio et~al.(2009)Bengio, Louradour, Collobert, and
  Weston}]{bengio2009curriculum}
Yoshua Bengio, J{\'e}r{\^o}me Louradour, Ronan Collobert, and Jason Weston.
  2009.
\newblock Curriculum learning.
\newblock In \emph{Proceedings of the 26th annual international conference on
  machine learning}, pages 41--48.

\bibitem[{Bowman et~al.(2015)Bowman, Angeli, Potts, and
  Manning}]{bowman-etal-2015-large}
Samuel~R. Bowman, Gabor Angeli, Christopher Potts, and Christopher~D. Manning.
  2015.
\newblock \href {https://doi.org/10.18653/v1/D15-1075} {A large annotated
  corpus for learning natural language inference}.
\newblock In \emph{Proceedings of the 2015 Conference on Empirical Methods in
  Natural Language Processing}, pages 632--642, Lisbon, Portugal. Association
  for Computational Linguistics.

\bibitem[{Cheng and Amiri(2024)}]{cheng2024mu}
Jiali Cheng and Hadi Amiri. 2024.
\newblock Mu-bench: A multitask multimodal benchmark for machine unlearning.
\newblock \emph{arXiv preprint arXiv:2406.14796}.

\bibitem[{Cheng et~al.(2024)Cheng, Elgaar, Vakil, and
  Amiri}]{cheng24c_interspeech}
Jiali Cheng, Mohamed Elgaar, Nidhi Vakil, and Hadi Amiri. 2024.
\newblock \href {https://doi.org/10.21437/Interspeech.2024-2370} {Cognivoice:
  Multimodal and multilingual fusion networks for mild cognitive impairment
  assessment from spontaneous speech}.
\newblock In \emph{Interspeech 2024}, pages 4308--4312.

\bibitem[{Cheng et~al.(2021)Cheng, Hao, Yuan, Si, and Carin}]{cheng2021fairfil}
Pengyu Cheng, Weituo Hao, Siyang Yuan, Shijing Si, and Lawrence Carin. 2021.
\newblock \href {https://openreview.net/forum?id=N6JECD-PI5w} {Fairfil:
  Contrastive neural debiasing method for pretrained text encoders}.
\newblock In \emph{9th International Conference on Learning Representations,
  {ICLR} 2021, Virtual Event, Austria, May 3-7, 2021}. OpenReview.net.

\bibitem[{Clark et~al.(2019)Clark, Yatskar, and
  Zettlemoyer}]{clark-etal-2019-dont}
Christopher Clark, Mark Yatskar, and Luke Zettlemoyer. 2019.
\newblock \href {https://doi.org/10.18653/v1/D19-1418} {Don{'}t take the easy
  way out: Ensemble based methods for avoiding known dataset biases}.
\newblock In \emph{Proceedings of the 2019 Conference on Empirical Methods in
  Natural Language Processing and the 9th International Joint Conference on
  Natural Language Processing (EMNLP-IJCNLP)}, pages 4069--4082, Hong Kong,
  China. Association for Computational Linguistics.

\bibitem[{Devlin et~al.(2019)Devlin, Chang, Lee, and
  Toutanova}]{devlin-etal-2019-bert}
Jacob Devlin, Ming-Wei Chang, Kenton Lee, and Kristina Toutanova. 2019.
\newblock \href {https://doi.org/10.18653/v1/N19-1423} {{BERT}: Pre-training of
  deep bidirectional transformers for language understanding}.
\newblock In \emph{Proceedings of the 2019 Conference of the North {A}merican
  Chapter of the Association for Computational Linguistics: Human Language
  Technologies, Volume 1 (Long and Short Papers)}, pages 4171--4186,
  Minneapolis, Minnesota. Association for Computational Linguistics.

\bibitem[{Dinan et~al.(2020)Dinan, Fan, Williams, Urbanek, Kiela, and
  Weston}]{dinan-etal-2020-queens}
Emily Dinan, Angela Fan, Adina Williams, Jack Urbanek, Douwe Kiela, and Jason
  Weston. 2020.
\newblock \href {https://doi.org/10.18653/v1/2020.emnlp-main.656} {Queens are
  powerful too: Mitigating gender bias in dialogue generation}.
\newblock In \emph{Proceedings of the 2020 Conference on Empirical Methods in
  Natural Language Processing (EMNLP)}, pages 8173--8188, Online. Association
  for Computational Linguistics.

\bibitem[{Dolan and Brockett(2005)}]{dolan-brockett-2005-automatically}
William~B. Dolan and Chris Brockett. 2005.
\newblock \href {https://aclanthology.org/I05-5002} {Automatically constructing
  a corpus of sentential paraphrases}.
\newblock In \emph{Proceedings of the Third International Workshop on
  Paraphrasing ({IWP}2005)}.

\bibitem[{Du et~al.(2023)Du, Ding, Sun, Liu, Qin, and
  Liu}]{du-etal-2023-towards}
Li~Du, Xiao Ding, Zhouhao Sun, Ting Liu, Bing Qin, and Jingshuo Liu. 2023.
\newblock \href {https://doi.org/10.18653/v1/2023.acl-long.161} {Towards stable
  natural language understanding via information entropy guided debiasing}.
\newblock In \emph{Proceedings of the 61st Annual Meeting of the Association
  for Computational Linguistics (Volume 1: Long Papers)}, pages 2868--2882,
  Toronto, Canada. Association for Computational Linguistics.

\bibitem[{Gandikota et~al.(2023)Gandikota, Materzy\'nska, Fiotto-Kaufman, and
  Bau}]{gandikota2023erasing}
Rohit Gandikota, Joanna Materzy\'nska, Jaden Fiotto-Kaufman, and David Bau.
  2023.
\newblock Erasing concepts from diffusion models.
\newblock In \emph{Proceedings of the 2023 IEEE International Conference on
  Computer Vision}.

\bibitem[{Gao et~al.(2022)Gao, Dou, Zhang, and Huang}]{gao-etal-2022-kernel}
SongYang Gao, Shihan Dou, Qi~Zhang, and Xuanjing Huang. 2022.
\newblock \href {https://doi.org/10.18653/v1/2022.emnlp-main.275}
  {Kernel-whitening: Overcome dataset bias with isotropic sentence embedding}.
\newblock In \emph{Proceedings of the 2022 Conference on Empirical Methods in
  Natural Language Processing}, pages 4112--4122, Abu Dhabi, United Arab
  Emirates. Association for Computational Linguistics.

\bibitem[{Gardner et~al.(2021)Gardner, Merrill, Dodge, Peters, Ross, Singh, and
  Smith}]{gardner-etal-2021-competency}
Matt Gardner, William Merrill, Jesse Dodge, Matthew Peters, Alexis Ross, Sameer
  Singh, and Noah~A. Smith. 2021.
\newblock \href {https://doi.org/10.18653/v1/2021.emnlp-main.135} {Competency
  problems: On finding and removing artifacts in language data}.
\newblock In \emph{Proceedings of the 2021 Conference on Empirical Methods in
  Natural Language Processing}, pages 1801--1813, Online and Punta Cana,
  Dominican Republic. Association for Computational Linguistics.

\bibitem[{Ghaddar et~al.(2021)Ghaddar, Langlais, Rezagholizadeh, and
  Rashid}]{ghaddar-etal-2021-end}
Abbas Ghaddar, Phillippe Langlais, Mehdi Rezagholizadeh, and Ahmad Rashid.
  2021.
\newblock \href {https://doi.org/10.18653/v1/2021.findings-acl.168} {End-to-end
  self-debiasing framework for robust {NLU} training}.
\newblock In \emph{Findings of the Association for Computational Linguistics:
  ACL-IJCNLP 2021}, pages 1923--1929, Online. Association for Computational
  Linguistics.

\bibitem[{Guo et~al.(2023)Guo, Tang, Ouyang, Wu, and
  Dai}]{guo-etal-2023-debias}
Qi~Guo, Yuanhang Tang, Yawen Ouyang, Zhen Wu, and Xinyu Dai. 2023.
\newblock \href {https://doi.org/10.18653/v1/2023.findings-emnlp.726} {Debias
  {NLU} datasets via training-free perturbations}.
\newblock In \emph{Findings of the Association for Computational Linguistics:
  EMNLP 2023}, pages 10886--10901, Singapore. Association for Computational
  Linguistics.

\bibitem[{Gururangan et~al.(2018)Gururangan, Swayamdipta, Levy, Schwartz,
  Bowman, and Smith}]{gururangan-etal-2018-annotation}
Suchin Gururangan, Swabha Swayamdipta, Omer Levy, Roy Schwartz, Samuel Bowman,
  and Noah~A. Smith. 2018.
\newblock \href {https://doi.org/10.18653/v1/N18-2017} {Annotation artifacts in
  natural language inference data}.
\newblock In \emph{Proceedings of the 2018 Conference of the North {A}merican
  Chapter of the Association for Computational Linguistics: Human Language
  Technologies, Volume 2 (Short Papers)}, pages 107--112, New Orleans,
  Louisiana. Association for Computational Linguistics.

\bibitem[{Gutmann and Hyvärinen(2010)}]{pmlr-v9-gutmann10a}
Michael Gutmann and Aapo Hyvärinen. 2010.
\newblock \href {https://proceedings.mlr.press/v9/gutmann10a.html}
  {Noise-contrastive estimation: A new estimation principle for unnormalized
  statistical models}.
\newblock In \emph{Proceedings of the Thirteenth International Conference on
  Artificial Intelligence and Statistics}, volume~9 of \emph{Proceedings of
  Machine Learning Research}, pages 297--304, Chia Laguna Resort, Sardinia,
  Italy. PMLR.

\bibitem[{Han et~al.(2022)Han, Baldwin, and Cohn}]{han-etal-2022-balancing}
Xudong Han, Timothy Baldwin, and Trevor Cohn. 2022.
\newblock \href {https://doi.org/10.18653/v1/2022.emnlp-main.779} {Balancing
  out bias: Achieving fairness through balanced training}.
\newblock In \emph{Proceedings of the 2022 Conference on Empirical Methods in
  Natural Language Processing}, pages 11335--11350, Abu Dhabi, United Arab
  Emirates. Association for Computational Linguistics.

\bibitem[{Hartvigsen et~al.(2022)Hartvigsen, Gabriel, Palangi, Sap, Ray, and
  Kamar}]{hartvigsen-etal-2022-toxigen}
Thomas Hartvigsen, Saadia Gabriel, Hamid Palangi, Maarten Sap, Dipankar Ray,
  and Ece Kamar. 2022.
\newblock \href {https://doi.org/10.18653/v1/2022.acl-long.234} {{T}oxi{G}en: A
  large-scale machine-generated dataset for adversarial and implicit hate
  speech detection}.
\newblock In \emph{Proceedings of the 60th Annual Meeting of the Association
  for Computational Linguistics (Volume 1: Long Papers)}, pages 3309--3326,
  Dublin, Ireland. Association for Computational Linguistics.

\bibitem[{Hinton(2002)}]{10.1162/089976602760128018}
Geoffrey~E. Hinton. 2002.
\newblock \href {https://doi.org/10.1162/089976602760128018} {Training products
  of experts by minimizing contrastive divergence}.
\newblock \emph{Neural Comput.}, 14(8):1771–1800.

\bibitem[{Jeon et~al.(2023)Jeon, Lee, Park, Kim, Mok, and
  Lee}]{jeon-etal-2023-improving}
Eojin Jeon, Mingyu Lee, Juhyeong Park, Yeachan Kim, Wing-Lam Mok, and SangKeun
  Lee. 2023.
\newblock \href {https://doi.org/10.18653/v1/2023.emnlp-main.681} {Improving
  bias mitigation through bias experts in natural language understanding}.
\newblock In \emph{Proceedings of the 2023 Conference on Empirical Methods in
  Natural Language Processing}, pages 11053--11066, Singapore. Association for
  Computational Linguistics.

\bibitem[{Jin et~al.(2021)Jin, Barbieri, Kennedy, Mostafazadeh~Davani, Neves,
  and Ren}]{jin-etal-2021-transferability}
Xisen Jin, Francesco Barbieri, Brendan Kennedy, Aida Mostafazadeh~Davani,
  Leonardo Neves, and Xiang Ren. 2021.
\newblock \href {https://doi.org/10.18653/v1/2021.naacl-main.296} {On
  transferability of bias mitigation effects in language model fine-tuning}.
\newblock In \emph{Proceedings of the 2021 Conference of the North American
  Chapter of the Association for Computational Linguistics: Human Language
  Technologies}, pages 3770--3783, Online. Association for Computational
  Linguistics.

\bibitem[{Karimi~Mahabadi et~al.(2020)Karimi~Mahabadi, Belinkov, and
  Henderson}]{karimi-mahabadi-etal-2020-end}
Rabeeh Karimi~Mahabadi, Yonatan Belinkov, and James Henderson. 2020.
\newblock \href {https://doi.org/10.18653/v1/2020.acl-main.769} {End-to-end
  bias mitigation by modelling biases in corpora}.
\newblock In \emph{Proceedings of the 58th Annual Meeting of the Association
  for Computational Linguistics}, pages 8706--8716, Online. Association for
  Computational Linguistics.

\bibitem[{Khosla et~al.(2020)Khosla, Teterwak, Wang, Sarna, Tian, Isola,
  Maschinot, Liu, and Krishnan}]{khosla2020supervised}
Prannay Khosla, Piotr Teterwak, Chen Wang, Aaron Sarna, Yonglong Tian, Phillip
  Isola, Aaron Maschinot, Ce~Liu, and Dilip Krishnan. 2020.
\newblock \href
  {https://proceedings.neurips.cc/paper/2020/hash/d89a66c7c80a29b1bdbab0f2a1a94af8-Abstract.html}
  {Supervised contrastive learning}.
\newblock In \emph{Advances in Neural Information Processing Systems 33: Annual
  Conference on Neural Information Processing Systems 2020, NeurIPS 2020,
  December 6-12, 2020, virtual}.

\bibitem[{Kim et~al.(2022)Kim, Hwang, Ahn, Park, and Kwak}]{kim2022learning}
Nayeong Kim, Sehyun Hwang, Sungsoo Ahn, Jaesik Park, and Suha Kwak. 2022.
\newblock \href {https://openreview.net/forum?id=nOw2HiKmvk1} {Learning
  debiased classifier with biased committee}.
\newblock In \emph{Advances in Neural Information Processing Systems}.

\bibitem[{Kingma and Ba(2015)}]{kingma2014adam}
Diederik~P. Kingma and Jimmy Ba. 2015.
\newblock \href {http://arxiv.org/abs/1412.6980} {Adam: {A} method for
  stochastic optimization}.
\newblock In \emph{3rd International Conference on Learning Representations,
  {ICLR} 2015, San Diego, CA, USA, May 7-9, 2015, Conference Track
  Proceedings}.

\bibitem[{Liu et~al.(2024)Liu, Dong, and Zhang}]{Liu_2024_WACV}
Xiulong Liu, Zhikang Dong, and Peng Zhang. 2024.
\newblock Tackling data bias in music-avqa: Crafting a balanced dataset for
  unbiased question-answering.
\newblock In \emph{Proceedings of the IEEE/CVF Winter Conference on
  Applications of Computer Vision (WACV)}, pages 4478--4487.

\bibitem[{Liu et~al.(2019)Liu, Ott, Goyal, Du, Joshi, Chen, Levy, Lewis,
  Zettlemoyer, and Stoyanov}]{liu2019roberta}
Yinhan Liu, Myle Ott, Naman Goyal, Jingfei Du, Mandar Joshi, Danqi Chen, Omer
  Levy, Mike Lewis, Luke Zettlemoyer, and Veselin Stoyanov. 2019.
\newblock \href {https://arxiv.org/abs/1907.11692} {Roberta: A robustly
  optimized bert pretraining approach}.
\newblock \emph{ArXiv preprint}, abs/1907.11692.

\bibitem[{Lyu et~al.(2022)Lyu, Li, Yang, de~Rijke, Ren, Zhao, Yin, and
  Ren}]{lyu2023feature}
Yougang Lyu, Piji Li, Yechang Yang, Maarten de~Rijke, Pengjie Ren, Yukun Zhao,
  Dawei Yin, and Zhaochun Ren. 2022.
\newblock Feature-level debiased natural language understanding.
\newblock In \emph{Proceedings of the AAAI Conference on Artificial
  Intelligence}.

\bibitem[{Madanagopal and Caverlee(2023)}]{madanagopal-caverlee-2023-bias}
Karthic Madanagopal and James Caverlee. 2023.
\newblock \href {https://doi.org/10.18653/v1/2023.emnlp-main.882} {Bias
  neutralization in non-parallel texts: A cyclic approach with auxiliary
  guidance}.
\newblock In \emph{Proceedings of the 2023 Conference on Empirical Methods in
  Natural Language Processing}, pages 14265--14278, Singapore. Association for
  Computational Linguistics.

\bibitem[{McCoy et~al.(2019)McCoy, Pavlick, and Linzen}]{mccoy-etal-2019-right}
Tom McCoy, Ellie Pavlick, and Tal Linzen. 2019.
\newblock \href {https://doi.org/10.18653/v1/P19-1334} {Right for the wrong
  reasons: Diagnosing syntactic heuristics in natural language inference}.
\newblock In \emph{Proceedings of the 57th Annual Meeting of the Association
  for Computational Linguistics}, pages 3428--3448, Florence, Italy.
  Association for Computational Linguistics.

\bibitem[{Meade et~al.(2022)Meade, Poole-Dayan, and
  Reddy}]{meade-etal-2022-empirical}
Nicholas Meade, Elinor Poole-Dayan, and Siva Reddy. 2022.
\newblock \href {https://doi.org/10.18653/v1/2022.acl-long.132} {An empirical
  survey of the effectiveness of debiasing techniques for pre-trained language
  models}.
\newblock In \emph{Proceedings of the 60th Annual Meeting of the Association
  for Computational Linguistics (Volume 1: Long Papers)}, pages 1878--1898,
  Dublin, Ireland. Association for Computational Linguistics.

\bibitem[{Meissner et~al.(2022)Meissner, Sugawara, and
  Aizawa}]{meissner-etal-2022-debiasing}
Johannes~Mario Meissner, Saku Sugawara, and Akiko Aizawa. 2022.
\newblock \href {https://doi.org/10.18653/v1/2022.emnlp-main.517} {Debiasing
  masks: A new framework for shortcut mitigation in {NLU}}.
\newblock In \emph{Proceedings of the 2022 Conference on Empirical Methods in
  Natural Language Processing}, pages 7607--7613, Abu Dhabi, United Arab
  Emirates. Association for Computational Linguistics.

\bibitem[{Mendelson and Belinkov(2021)}]{mendelson-belinkov-2021-debiasing}
Michael Mendelson and Yonatan Belinkov. 2021.
\newblock \href {https://doi.org/10.18653/v1/2021.emnlp-main.116} {Debiasing
  methods in natural language understanding make bias more accessible}.
\newblock In \emph{Proceedings of the 2021 Conference on Empirical Methods in
  Natural Language Processing}, pages 1545--1557, Online and Punta Cana,
  Dominican Republic. Association for Computational Linguistics.

\bibitem[{Modarressi et~al.(2023)Modarressi, Amirkhani, and
  Pilehvar}]{modarressi-etal-2023-guide}
Ali Modarressi, Hossein Amirkhani, and Mohammad~Taher Pilehvar. 2023.
\newblock \href {https://doi.org/10.18653/v1/2023.eacl-main.143} {Guide the
  learner: Controlling product of experts debiasing method based on token
  attribution similarities}.
\newblock In \emph{Proceedings of the 17th Conference of the European Chapter
  of the Association for Computational Linguistics}, pages 1954--1959,
  Dubrovnik, Croatia. Association for Computational Linguistics.

\bibitem[{Nadeem et~al.(2021)Nadeem, Bethke, and
  Reddy}]{nadeem-etal-2021-stereoset}
Moin Nadeem, Anna Bethke, and Siva Reddy. 2021.
\newblock \href {https://doi.org/10.18653/v1/2021.acl-long.416} {{S}tereo{S}et:
  Measuring stereotypical bias in pretrained language models}.
\newblock In \emph{Proceedings of the 59th Annual Meeting of the Association
  for Computational Linguistics and the 11th International Joint Conference on
  Natural Language Processing (Volume 1: Long Papers)}, pages 5356--5371,
  Online. Association for Computational Linguistics.

\bibitem[{Naik et~al.(2018)Naik, Ravichander, Sadeh, Rose, and
  Neubig}]{naik-etal-2018-stress}
Aakanksha Naik, Abhilasha Ravichander, Norman Sadeh, Carolyn Rose, and Graham
  Neubig. 2018.
\newblock \href {https://aclanthology.org/C18-1198} {Stress test evaluation for
  natural language inference}.
\newblock In \emph{Proceedings of the 27th International Conference on
  Computational Linguistics}, pages 2340--2353, Santa Fe, New Mexico, USA.
  Association for Computational Linguistics.

\bibitem[{Poliak et~al.(2018)Poliak, Naradowsky, Haldar, Rudinger, and
  Van~Durme}]{poliak-etal-2018-hypothesis}
Adam Poliak, Jason Naradowsky, Aparajita Haldar, Rachel Rudinger, and Benjamin
  Van~Durme. 2018.
\newblock \href {https://doi.org/10.18653/v1/S18-2023} {Hypothesis only
  baselines in natural language inference}.
\newblock In \emph{Proceedings of the Seventh Joint Conference on Lexical and
  Computational Semantics}, pages 180--191, New Orleans, Louisiana. Association
  for Computational Linguistics.

\bibitem[{Pool and Yu(2021)}]{NEURIPS2021_6e8404c3}
Jeff Pool and Chong Yu. 2021.
\newblock \href
  {https://proceedings.neurips.cc/paper/2021/hash/6e8404c3b93a9527c8db241a1846599a-Abstract.html}
  {Channel permutations for {N:} {M} sparsity}.
\newblock In \emph{Advances in Neural Information Processing Systems 34: Annual
  Conference on Neural Information Processing Systems 2021, NeurIPS 2021,
  December 6-14, 2021, virtual}, pages 13316--13327.

\bibitem[{Qian et~al.(2021)Qian, Feng, Wen, Ma, and
  Xie}]{qian-etal-2021-counterfactual}
Chen Qian, Fuli Feng, Lijie Wen, Chunping Ma, and Pengjun Xie. 2021.
\newblock \href {https://doi.org/10.18653/v1/2021.acl-long.422} {Counterfactual
  inference for text classification debiasing}.
\newblock In \emph{Proceedings of the 59th Annual Meeting of the Association
  for Computational Linguistics and the 11th International Joint Conference on
  Natural Language Processing (Volume 1: Long Papers)}, pages 5434--5445,
  Online. Association for Computational Linguistics.

\bibitem[{Radford et~al.(2019)Radford, Wu, Child, Luan, Amodei, Sutskever
  et~al.}]{radford2019language}
Alec Radford, Jeffrey Wu, Rewon Child, David Luan, Dario Amodei, Ilya
  Sutskever, et~al. 2019.
\newblock Language models are unsupervised multitask learners.
\newblock \emph{OpenAI blog}, 1(8):9.

\bibitem[{Ravichander et~al.(2023)Ravichander, Stacey, and
  Rei}]{ravichander-etal-2023-bias}
Abhilasha Ravichander, Joe Stacey, and Marek Rei. 2023.
\newblock \href {https://doi.org/10.18653/v1/2023.findings-emnlp.619} {When and
  why does bias mitigation work?}
\newblock In \emph{Findings of the Association for Computational Linguistics:
  EMNLP 2023}, pages 9233--9247, Singapore. Association for Computational
  Linguistics.

\bibitem[{Reif and Schwartz(2023)}]{reif-schwartz-2023-fighting}
Yuval Reif and Roy Schwartz. 2023.
\newblock \href {https://doi.org/10.18653/v1/2023.findings-acl.833} {Fighting
  bias with bias: Promoting model robustness by amplifying dataset biases}.
\newblock In \emph{Findings of the Association for Computational Linguistics:
  ACL 2023}, pages 13169--13189, Toronto, Canada. Association for Computational
  Linguistics.

\bibitem[{Sanh et~al.(2021)Sanh, Wolf, Belinkov, and Rush}]{sanh2020learning}
Victor Sanh, Thomas Wolf, Yonatan Belinkov, and Alexander~M. Rush. 2021.
\newblock \href {https://openreview.net/forum?id=Hf3qXoiNkR} {Learning from
  others' mistakes: Avoiding dataset biases without modeling them}.
\newblock In \emph{9th International Conference on Learning Representations,
  {ICLR} 2021, Virtual Event, Austria, May 3-7, 2021}. OpenReview.net.

\bibitem[{Schick et~al.(2021)Schick, Udupa, and
  Sch{\"u}tze}]{schick-etal-2021-self}
Timo Schick, Sahana Udupa, and Hinrich Sch{\"u}tze. 2021.
\newblock \href {https://doi.org/10.1162/tacl_a_00434} {Self-diagnosis and
  self-debiasing: A proposal for reducing corpus-based bias in {NLP}}.
\newblock \emph{Transactions of the Association for Computational Linguistics},
  9:1408--1424.

\bibitem[{Sharma et~al.(2019)Sharma, Graesser, Nangia, and
  Evci}]{sharma2019natural}
Lakshay Sharma, Laura Graesser, Nikita Nangia, and Utku Evci. 2019.
\newblock \href {https://arxiv.org/abs/1907.01041} {Natural language
  understanding with the quora question pairs dataset}.
\newblock \emph{ArXiv preprint}, abs/1907.01041.

\bibitem[{Shen et~al.(2022)Shen, Han, Cohn, Baldwin, and
  Frermann}]{shen-etal-2022-representational}
Aili Shen, Xudong Han, Trevor Cohn, Timothy Baldwin, and Lea Frermann. 2022.
\newblock \href {https://aclanthology.org/2022.findings-aacl.8} {Does
  representational fairness imply empirical fairness?}
\newblock In \emph{Findings of the Association for Computational Linguistics:
  AACL-IJCNLP 2022}, pages 81--95, Online only. Association for Computational
  Linguistics.

\bibitem[{Sinha et~al.(2021)Sinha, Parthasarathi, Pineau, and
  Williams}]{sinha-etal-2021-unnatural}
Koustuv Sinha, Prasanna Parthasarathi, Joelle Pineau, and Adina Williams. 2021.
\newblock \href {https://doi.org/10.18653/v1/2021.acl-long.569} {{UnNatural}
  {L}anguage {I}nference}.
\newblock In \emph{Proceedings of the 59th Annual Meeting of the Association
  for Computational Linguistics and the 11th International Joint Conference on
  Natural Language Processing (Volume 1: Long Papers)}, pages 7329--7346,
  Online. Association for Computational Linguistics.

\bibitem[{Sohn(2016)}]{NIPS2016_6b180037}
Kihyuk Sohn. 2016.
\newblock \href
  {https://proceedings.neurips.cc/paper/2016/hash/6b180037abbebea991d8b1232f8a8ca9-Abstract.html}
  {Improved deep metric learning with multi-class n-pair loss objective}.
\newblock In \emph{Advances in Neural Information Processing Systems 29: Annual
  Conference on Neural Information Processing Systems 2016, December 5-10,
  2016, Barcelona, Spain}, pages 1849--1857.

\bibitem[{Sousa et~al.(2019)Sousa, Lamurias, and
  Couto}]{sousa-etal-2019-silver}
Diana Sousa, Andre Lamurias, and Francisco~M. Couto. 2019.
\newblock \href {https://doi.org/10.18653/v1/N19-1152} {A silver standard
  corpus of human phenotype-gene relations}.
\newblock In \emph{Proceedings of the 2019 Conference of the North {A}merican
  Chapter of the Association for Computational Linguistics: Human Language
  Technologies, Volume 1 (Long and Short Papers)}, pages 1487--1492,
  Minneapolis, Minnesota. Association for Computational Linguistics.

\bibitem[{Sun et~al.(2022)Sun, Thai, and Iyyer}]{sun-etal-2022-chapterbreak}
Simeng Sun, Katherine Thai, and Mohit Iyyer. 2022.
\newblock \href {https://doi.org/10.18653/v1/2022.naacl-main.271}
  {{C}hapter{B}reak: A challenge dataset for long-range language models}.
\newblock In \emph{Proceedings of the 2022 Conference of the North American
  Chapter of the Association for Computational Linguistics: Human Language
  Technologies}, pages 3704--3714, Seattle, United States. Association for
  Computational Linguistics.

\bibitem[{Tsirigotis et~al.(2023)Tsirigotis, Monteiro, Rodriguez, Vazquez, and
  Courville}]{NEURIPS2023_b0d9ceb3}
Christos Tsirigotis, Joao Monteiro, Pau Rodriguez, David Vazquez, and Aaron~C
  Courville. 2023.
\newblock \href
  {https://proceedings.neurips.cc/paper_files/paper/2023/file/b0d9ceb3d11d013e55da201d2a2c07b2-Paper-Conference.pdf}
  {Group robust classification without any group information}.
\newblock In \emph{Advances in Neural Information Processing Systems},
  volume~36, pages 56553--56575. Curran Associates, Inc.

\bibitem[{Utama et~al.(2020{\natexlab{a}})Utama, Moosavi, and
  Gurevych}]{utama-etal-2020-mind}
Prasetya~Ajie Utama, Nafise~Sadat Moosavi, and Iryna Gurevych.
  2020{\natexlab{a}}.
\newblock \href {https://doi.org/10.18653/v1/2020.acl-main.770} {Mind the
  trade-off: Debiasing {NLU} models without degrading the in-distribution
  performance}.
\newblock In \emph{Proceedings of the 58th Annual Meeting of the Association
  for Computational Linguistics}, pages 8717--8729, Online. Association for
  Computational Linguistics.

\bibitem[{Utama et~al.(2020{\natexlab{b}})Utama, Moosavi, and
  Gurevych}]{utama-etal-2020-towards}
Prasetya~Ajie Utama, Nafise~Sadat Moosavi, and Iryna Gurevych.
  2020{\natexlab{b}}.
\newblock \href {https://doi.org/10.18653/v1/2020.emnlp-main.613} {Towards
  debiasing {NLU} models from unknown biases}.
\newblock In \emph{Proceedings of the 2020 Conference on Empirical Methods in
  Natural Language Processing (EMNLP)}, pages 7597--7610, Online. Association
  for Computational Linguistics.

\bibitem[{Vakil and Amiri(2022)}]{vakil-amiri-2022-generic}
Nidhi Vakil and Hadi Amiri. 2022.
\newblock \href {https://doi.org/10.18653/v1/2022.naacl-main.160} {Generic and
  trend-aware curriculum learning for relation extraction}.
\newblock In \emph{Proceedings of the 2022 Conference of the North American
  Chapter of the Association for Computational Linguistics: Human Language
  Technologies}, pages 2202--2213, Seattle, United States. Association for
  Computational Linguistics.

\bibitem[{Wang et~al.(2023)Wang, Huang, Yan, Zhou, and
  Chen}]{wang-etal-2023-robust}
Fei Wang, James~Y. Huang, Tianyi Yan, Wenxuan Zhou, and Muhao Chen. 2023.
\newblock \href {https://doi.org/10.18653/v1/2023.findings-acl.32} {Robust
  natural language understanding with residual attention debiasing}.
\newblock In \emph{Findings of the Association for Computational Linguistics:
  ACL 2023}, pages 504--519, Toronto, Canada. Association for Computational
  Linguistics.

\bibitem[{Williams et~al.(2018)Williams, Nangia, and
  Bowman}]{williams-etal-2018-broad}
Adina Williams, Nikita Nangia, and Samuel Bowman. 2018.
\newblock \href {https://doi.org/10.18653/v1/N18-1101} {A broad-coverage
  challenge corpus for sentence understanding through inference}.
\newblock In \emph{Proceedings of the 2018 Conference of the North {A}merican
  Chapter of the Association for Computational Linguistics: Human Language
  Technologies, Volume 1 (Long Papers)}, pages 1112--1122, New Orleans,
  Louisiana. Association for Computational Linguistics.

\bibitem[{Wu et~al.(2022)Wu, Gardner, Stenetorp, and
  Dasigi}]{wu-etal-2022-generating}
Yuxiang Wu, Matt Gardner, Pontus Stenetorp, and Pradeep Dasigi. 2022.
\newblock \href {https://doi.org/10.18653/v1/2022.acl-long.190} {Generating
  data to mitigate spurious correlations in natural language inference
  datasets}.
\newblock In \emph{Proceedings of the 60th Annual Meeting of the Association
  for Computational Linguistics (Volume 1: Long Papers)}, pages 2660--2676,
  Dublin, Ireland. Association for Computational Linguistics.

\bibitem[{Xu et~al.(2015)Xu, Jin, Shen, and Zhu}]{Xu_Jin_Shen_Zhu_2015}
Zenglin Xu, Rong Jin, Bin Shen, and Shenghuo Zhu. 2015.
\newblock \href {https://doi.org/10.1609/aaai.v29i1.9626} {Nystrom
  approximation for sparse kernel methods: Theoretical analysis and empirical
  evaluation}.
\newblock \emph{Proceedings of the AAAI Conference on Artificial Intelligence},
  29(1).

\bibitem[{Yao et~al.(2023)Yao, Xu, and Liu}]{yao2023large}
Yuanshun Yao, Xiaojun Xu, and Yang Liu. 2023.
\newblock Large language model unlearning.
\newblock \emph{arXiv preprint arXiv:2310.10683}.

\bibitem[{Yu et~al.(2023)Yu, Jeoung, Kasi, Yu, and
  Ji}]{yu-etal-2023-unlearning}
Charles Yu, Sullam Jeoung, Anish Kasi, Pengfei Yu, and Heng Ji. 2023.
\newblock \href {https://doi.org/10.18653/v1/2023.findings-acl.375} {Unlearning
  bias in language models by partitioning gradients}.
\newblock In \emph{Findings of the Association for Computational Linguistics:
  ACL 2023}, pages 6032--6048, Toronto, Canada. Association for Computational
  Linguistics.

\bibitem[{Zhang et~al.(2019)Zhang, Baldridge, and He}]{zhang-etal-2019-paws}
Yuan Zhang, Jason Baldridge, and Luheng He. 2019.
\newblock \href {https://doi.org/10.18653/v1/N19-1131} {{PAWS}: Paraphrase
  adversaries from word scrambling}.
\newblock In \emph{Proceedings of the 2019 Conference of the North {A}merican
  Chapter of the Association for Computational Linguistics: Human Language
  Technologies, Volume 1 (Long and Short Papers)}, pages 1298--1308,
  Minneapolis, Minnesota. Association for Computational Linguistics.

\bibitem[{Zmigrod et~al.(2019)Zmigrod, Mielke, Wallach, and
  Cotterell}]{zmigrod-etal-2019-counterfactual}
Ran Zmigrod, Sabrina~J. Mielke, Hanna Wallach, and Ryan Cotterell. 2019.
\newblock \href {https://doi.org/10.18653/v1/P19-1161} {Counterfactual data
  augmentation for mitigating gender stereotypes in languages with rich
  morphology}.
\newblock In \emph{Proceedings of the 57th Annual Meeting of the Association
  for Computational Linguistics}, pages 1651--1661, Florence, Italy.
  Association for Computational Linguistics.

\end{thebibliography}

\appendix

\section{Implementation Details}
For all datasets, we train all methods on the BERT-base~\citep{devlin-etal-2019-bert} checkpoint, with a 2e-5 learning rate with linear decay using AdamW~\citep{kingma2014adam} optimizer. The batch size is set to 32. For the baseline models, we follow their papers for the hyperparameter choices. All experiments on done on a single A100 GPU.


We implement the proposed perturbation as illustrated in Table~\ref{tab:aug_list} by randomly dropping $50\%$ of the tokens from each sentence, dropping all layers after the second layer (3--12), 
and zeroing $m=90\%$ of the elements in the intact representation $f(x_i)$. Each branch-specific MLP consists of two linear layers with a ReLU activation function in between. 
We use $\lambda = 0.1$ in our experiments.

\section{Other Debiasing Objectives}
\label{sec:debias_obj}
The idea of existing debiasing objectives is based on the idea of adjusting the importance of training examples, i.e. their contribution to loss calculation. The importance of examples which the model fails the correctly predict is promoted while the importance of examples which the model correctly predicts is reduced. 

Product-of-Experts (PoE)~\citep{clark-etal-2019-dont,karimi-mahabadi-etal-2020-end,sanh2020learning} is one of the most commonly adopted debiasing objective, which takes dot product of the logits of the main model and the biased models. Debiasing Focal Loss~\citep{karimi-mahabadi-etal-2020-end} down-weights the main model based on how close the logits of the biased models is to 1. Confidence Regularization~\citep{utama-etal-2020-towards} reduced the loss scale of examples with a scaling mechanism.

\section{RoBERTa as Encoder}
We conducted experiments on RoBERTa-base~\citep{liu2019roberta} using the MNLI dataset to evaluate the efficacy of FairFlow more effectively. The results in Table~\ref{tab:roberta} shows that the performance of all models improved using RoBERTa-base as encoder. We also observe comparable gains to BERT as encoder in case of ID and Transfer settings and smaller gains in case of Stress and OOD settings, which can be attributed to the use of a more powerful encoder.

\begin{table*}
\small
\centering
\begin{tabular}{l|cccc|cccc|cc}
\toprule
\textbf{Model} & \multicolumn{4}{c|}{\textbf{MNLI} (Acc.)} & \multicolumn{4}{c|}{\textbf{QQP} (F1)} & \multicolumn{2}{c}{\textbf{PGR} (F1)} \\
               & ID   & Stress & OOD & Transfer & ID & Stress & OOD & Transfer & ID & Stress \\ \toprule
    \textbf{\FT}        & 84.4 & 55.8 & 60.7 & 80.1 & 89.1 & 59.3 & 40.8 & 61.8 & 67.1 & 54.3 \\
    \midrule
    \textbf{\MASK}      & 84.7 & 53.6 & 60.8 & 80.5 & 88.3 & 60.2 & 44.7 & 62.1 & 65.4 & 44.6 \\
    \textbf{\KW}        & 83.3 & 53.5 & 60.5 & 80.2 & 87.6 & 61.3 & 45.1 & 62.7 & 63.5 & 42.0 \\
    \textbf{\ETE}       & 83.8 & 57.8 & 66.3 & 80.1 & 89.2 & 58.9 & 42.5 & \underline{63.1} & 63.2 & 50.3 \\
    \textbf{LWBC}	    & 83.2 & 58.3 & 60.2 & 80.7 & 89.6 & 73.2 & 49.2 & 67.4 & 66.5 & 53.2 \\
    \textbf{\IE}        & 84.5 & 60.1 & 67.2 & 79.8 & 84.6 & 57.3 & \underline{50.6} & 60.5 & 64.8 & 54.6 \\
    \textbf{\READ}      & 79.6 & 58.3 & \textbf{68.4} & 73.0 & 84.5 & 65.8 & 46.7 & 61.7 & 62.6 & 55.0 \\
    \midrule
    \textbf{\OursPoe}   & \underline{84.8} & 62.3 & 67.5 & \underline{81.0} & 89.2 & 77.5 & 48.9 & \underline{63.1} & \underline{67.4} & 55.6 \\
    \textbf{\OursFocal} & \underline{84.8} & \underline{62.8} & \underline{67.9} & 80.9 & \underline{89.6} & \underline{77.8} & 49.2 & \underline{63.1} & \textbf{67.7} & \textbf{56.1} \\
    \textbf{\OursCL}    & \textbf{84.9}    & \textbf{63.6} & \textbf{68.4} & \textbf{81.1} & \textbf{91.8} & \textbf{78.4} & \textbf{51.5} & \textbf{68.3} & \textbf{67.7} & \underline{55.8} \\ \bottomrule
  \end{tabular}
  \caption{Experimental results on three datasets using BERT as the base model. The best performance is in \textbf{bold} and the second best is \underline{underlined}. Note that \IE does not release their code. We tried our best to reproduce the results but failed on HANS, which is 5.2 points lower than the reported 72.4. This is potentially due to implementation and optimization details which the authors did not release.}
  \label{tab:bert}
\end{table*}

\begin{table*}[ht]
\small
\centering
\begin{tabular}{l|cccc|cccc|cc}
\toprule
\multirow{2}{2em}{\textbf{Model}} & \multicolumn{4}{c|}{\textbf{MNLI} (Acc.)} & \multicolumn{4}{c|}{\textbf{QQP} (F1)} & \multicolumn{2}{c}{\textbf{PGR} (F1)} \\
               & ID   & Stress & OOD & Transfer & ID & Stress & OOD & Transfer & ID & Stress \\
    \toprule
    \textbf{\FT}        & 88.1 & 75.3 & 66.4 & 81.0 & 92.2 & 63.5 & 44.7 & 68.3 & 69.3 & 57.1 \\
    \midrule
    \textbf{\MASK}      & 86.5 & 72.7 & 66.9 & 80.7 & 92.5 & 66.1 & 49.1 & 68.7 & 70.2 & 57.5 \\
    \textbf{\KW}        & 88.1 & 74.1 & 67.4 & 79.9 & 93.1 & 66.7 & 50.2 & 68.9 & 71.3 & 58.3 \\
    \textbf{\ETE}       & 88.3 & 72.6 & 69.5 & 80.7 & 92.4 & 66.4 & 50.3 & 68.5 & 70.5 & 57.9 \\
    \textbf{LWBC}       & 84.6 & 69.3 & 66.7 & 81.0 & 91.7 & 63.2 & 43.9 & 67.4 & 70.4 & 54.2 \\
    \textbf{\IE}        & 88.2 & 72.4 & 69.3 & 80.3 & 92.3 & 66.3 & 50.2 & 68.3 & 70.8 & 56.3 \\
    \textbf{\READ}      & 85.3 & 73.5 & 70.3 & 78.5 & 91.4 & 68.1 & 51.0 & 67.8 & 69.3 & 55.7 \\
    \midrule
    \textbf{\OursPoe}   & 88.3 & 76.1 & 70.2 & 81.4 & 92.5 & 66.7 & 50.6 & 68.3 & 70.8 & 58.0 \\
    \textbf{\OursFocal} & 88.2 & 76.7 & 70.3 & 81.4 & 92.7 & 67.8 & 51.3 & 68.7 & 71.1 & 58.3 \\
    \textbf{\OursCL}    & 88.3 & 77.2 & 70.4 & 81.2 & 93.3 & 68.4 & 51.8 & 68.6 & 71.4 & 58.3 \\
    \bottomrule
  \end{tabular}
  \caption{Results using RoBERTa~\citep{liu2019roberta} as the base model. The best performance is in \textbf{bold} and the second best is \underline{underlined}.}
  \label{tab:roberta}
\end{table*}

\begin{table*}
\small
\centering
\begin{tabular}{l|cccc|cccc|cc}
\toprule
\multirow{2}{2em}{\textbf{Model}} & \multicolumn{4}{c|}{\textbf{MNLI} (Acc.)} & \multicolumn{4}{c|}{\textbf{QQP} (F1)} & \multicolumn{2}{c}{\textbf{PGR} (F1)} \\
               & ID   & Stress & OOD & Transfer & ID & Stress & OOD & Transfer & ID & Stress \\ \toprule
    \textbf{\FT}        & 80.4 & 54.0 & 52.1 & 75.2 & 84.5 & 67.3 & 57.7 & 65.1 & 56.5 & 54.2 \\
    \midrule
    \textbf{\MASK}      & 79.3 & 52.8 & 51.6 & 73.7 & 83.6 & 67.5 & 57.3 & 74.9 & 56.7 & 53.2 \\
    \textbf{\KW}        & 80.7 & 55.1 & 52.8 & 75.1 & 85.7 & 67.4 & 58.3 & 77.2 & 58.3 & 55.1 \\
    \textbf{\ETE}       & 78.1 & 53.6 & 51.1 & 71.9 & 83.9 & 68.2 & 61.5 & 80.1 & 55.4 & 52.7 \\
    \textbf{LWBC}       & 74.5 & 50.8 & 51.0 & 71.4 & 80.2 & 61.0 & 55.8 & 75.4 & 53.2 & 51.0 \\
    \textbf{\IE}        & 79.7 & 52.9 & 51.8 & 74.3 & 86.1 & 66.9 & 58.2 & 76.2 & 57.1 & 53.8 \\
    \textbf{\READ}      & 77.5 & 52.7 & 51.5 & 73.8 & 85.2 & 66.4 & 63.3 & 75.2 & 57.3 & 52.6 \\
    \midrule
    \textbf{\OursPoe}   & 80.9 & 54.7 & 55.2 & 76.2 & 84.7 & 68.8 & 62.3 & 80.0 & 56.5 & 54.1 \\
    \textbf{\OursFocal} & 81.8 & 55.1 & 54.9 & 75.8 & 86.3 & 68.5 & 64.2 & 80.4 & 57.4 & 55.3 \\
    \textbf{\OursCL}    & 82.2 & 55.6 & 56.1 & 76.7 & 86.3 & 69.3 & 64.7 & 80.4 & 58.6 & 55.8 \\
    \bottomrule
  \end{tabular}
  \caption{Results using GPT-2 as the base model. The best performance is in \textbf{bold} and the second best is \underline{underlined}.}
  \label{tab:gpt2}
\end{table*}

\begin{table}
\small
\centering
\begin{tabular}{l|cc}
\toprule
\textbf{Model} & \textbf{Avg. Acc ($\uparrow$)} & \textbf{Std. Acc ($\downarrow$)} \\
    \midrule
    \textbf{\FT}        & 60.1 & 9.3 \\ 
    \midrule
    \textbf{\MASK}      & 58.7 & 6.7 \\ 
    \textbf{\KW}        & 59.3 & 5.9 \\ 
    \textbf{\ETE}       & 60.0 & 6.1 \\
    \textbf{LSWC}       & 59.4 & 5.8 \\ 
    \textbf{\IE}        & 60.1 & 7.3 \\
    \textbf{\READ}      & 60.9 & 5.6 \\ 
    \midrule
    \textbf{\OursPoe}   & 63.8 & 5.7 \\ 
    \textbf{\OursFocal} & \underline{64.3} & \underline{5.2} \\
    \textbf{\OursCL}    & \textbf{64.7} & \textbf{5.1} \\ \bottomrule
  \end{tabular}
  \caption{Average performance and standard deviation on each type of stress test averaged across three architectures. The best performance is in \textbf{bold} and the second best is \underline{underlined}.}
  \label{tab:subset}
\end{table}

\section{Perturbation for Data Augmentation}
The explicit perturbation operators proposed in our framework offer a valuable opportunity for data augmentation. This can be particularly useful in tasks such as NLI. Consider the example $(x_i^p, x_i^h, y_i)$, where $x_i^p$ represents the premise, $x_i^h$ represents the hypothesis, and $y_i$ denotes the label. To augment the dataset, we create additional data samples by applying different perturbation operations, e.g., by dropping the premise: (`', $x_i^h$, not entailment), dropping the hypothesis: ($x_i^p$, `', not entailment), shuffling the data: ($\mathcal{P}_{Irr}(x_i^p)$, $\mathcal{P}_{Irr}(x_i^h)$, not entailment) and dropping parts of the input: ($\mathcal{P}_{Sub}(x_i^p)$, $\mathcal{P}_{Irr}(x_i^h)$, not entailment). 
The augmented examples can be added back to the original dataset to mitigate the effect of bias during fine-tuning and potentially enhance model's generalizability, leading to improved performance on existing debiasing methods (See Table~\ref{tab:res_aug}).

\begin{table*}[ht]
\small
\centering  
\begin{tabular}{l|p{.6cm}p{.6cm}p{.6cm}p{.9cm}}
\toprule
\multirow{2}{2em}{\textbf{Model}} & \multicolumn{4}{c}{\textbf{MNLI} (Acc.)} \\
               & ID   & Stress & OOD & Transfer \\
    \toprule
    \textbf{\FT}         & 84.4 & 55.8 & 60.7 & 80.1 \\
    \textbf{\FT + Aug}   & 84.5 & 59.1 & 61.0 & 81.0 \\
    \midrule
    \textbf{\MASK}       & 84.7 & 53.6 & 60.8 & 80.5 \\
    \textbf{\MASK + Aug} & 85.6 & 55.4 & 62.2 & 81.1 \\
    \textbf{\KW}         & 83.3 & 53.5 & 60.5 & 80.2 \\
    \textbf{\KW + Aug}   & 85.1 & 56.2 & 60.8 & 81.0 \\
    \textbf{\ETE}        & 83.8 & 57.8 & 66.3 & 80.1 \\
    \textbf{\ETE + Aug}  & 84.8 & 61.1 & 66.2 & 80.6 \\
    \textbf{\IE}         & 84.5 & 60.1 & 65.7 & 79.8 \\
    \textbf{\IE + Aug}   & 85.6 & 60.8 & 66.4 & 80.7 \\
    \textbf{\READ}       & 79.6 & 58.3 & 68.4 & 73.0 \\
    \textbf{\READ + Aug} & 79.6 & 58.3 & 69.6 & 77.2 \\
    \bottomrule
  \end{tabular}
  \caption{Performance when applying data augmentation, which effectively improve existing debiasing methods. The best performance is in \textbf{bold}.}
  \label{tab:res_aug}
\end{table*}

\end{document}